\documentclass[manuscript,screen,nonacm]{acmart}%

\AtBeginDocument{%
  \providecommand\BibTeX{{%
    \normalfont B\kern-0.5em{\scshape i\kern-0.25em b}\kern-0.8em\TeX}}}

\setcopyright{none}%

\usepackage{tasks}
\usepackage{todonotes}
\usepackage{wrapfig}

\usepackage{enumitem}

\usepackage{fontawesome}

\usepackage[framemethod=tikz]{mdframed}
\definecolor{mycolor}{rgb}{0.122, 0.435, 0.698}
\definecolor{myblue}{rgb}{0.122, 0.435, 0.698}
\definecolor{mygreen}{rgb}{0.125, 0.525, 0.220}
\definecolor{myyellow}{rgb}{0.588, 0.439, 0.000}
\definecolor{myred}{rgb}{0.647, 0.114, 0.165}
\newmdenv[innerlinewidth=0.5pt,roundcorner=4pt,innerleftmargin=6pt,
          innerrightmargin=6pt,innertopmargin=6pt,innerbottommargin=6pt,
          linecolor=mycolor,backgroundcolor=mycolor!25!white]{mybox}
\newmdenv[innerlinewidth=0.5pt,roundcorner=4pt,innerleftmargin=6pt,
          innerrightmargin=6pt,innertopmargin=6pt,innerbottommargin=6pt,
          linecolor=myblue,backgroundcolor=myblue!25!white]{mybluebox}
\newmdenv[innerlinewidth=0.5pt,roundcorner=4pt,innerleftmargin=6pt,
          innerrightmargin=6pt,innertopmargin=6pt,innerbottommargin=6pt,
          linecolor=mygreen,backgroundcolor=mygreen!25!white]{mygreenbox}
\newmdenv[innerlinewidth=0.5pt,roundcorner=4pt,innerleftmargin=6pt,
          innerrightmargin=6pt,innertopmargin=6pt,innerbottommargin=6pt,
          linecolor=myyellow,backgroundcolor=myyellow!25!white]{myyellowbox}
\newmdenv[innerlinewidth=0.5pt,roundcorner=4pt,innerleftmargin=6pt,
          innerrightmargin=6pt,innertopmargin=6pt,innerbottommargin=6pt,
          linecolor=myred,backgroundcolor=myred!25!white]{myredbox}

\newif\ifdisplayGuidelines
\displayGuidelinestrue
\displayGuidelinesfalse

\begin{document}

\title[%
Explainability Is in the Mind of the Beholder%
]{%
Explainability Is in the Mind of the Beholder:\\ %
Establishing the Foundations of Explainable Artificial Intelligence%
}

\author{Kacper Sokol}
\authornote{Corresponding author.}
\authornote{Lead contact.}
\email{K.Sokol@bristol.ac.uk}
\email{Kacper.Sokol@rmit.edu.au}
\orcid{0000-0002-9869-5896}
\affiliation{%
  \institution{Intelligent Systems Laboratory, University of Bristol}
  \country{United Kingdom}
}
\affiliation{%
  \institution{ARC Centre of Excellence for Automated Decision-Making and Society, RMIT University}
  \country{Australia}
}

\author{Peter Flach}
\email{Peter.Flach@bristol.ac.uk}
\orcid{0000-0001-6857-5810}
\affiliation{%
  \institution{Intelligent Systems Laboratory, University of Bristol}
  \country{United Kingdom}
}

\begin{abstract}
Explainable artificial intelligence and interpretable machine learning are research domains growing in importance. %
Yet, the underlying concepts remain somewhat elusive and lack generally agreed definitions. %
While recent inspiration from social sciences has refocused the work on needs and expectations of human recipients, the field still misses a concrete conceptualisation. %
We take steps towards addressing this challenge by reviewing the philosophical and social foundations of human explainability, which we then translate into the technological realm. %
In particular, we scrutinise the notion of algorithmic black boxes and the spectrum of understanding determined by explanatory processes and explainees' background knowledge. %
This approach allows us to define explainability as (logical) \emph{reasoning} applied to transparent \emph{insights} (into, possibly black-box, predictive systems) interpreted under \emph{background knowledge} and placed within a specific context -- a process that engenders \emph{understanding} in a selected group of explainees. %
We then employ this conceptualisation to revisit strategies for evaluating explainability as well as the much disputed trade-off between transparency and predictive power, including its implications for ante-hoc and post-hoc techniques along with fairness and accountability established by explainability. %
We furthermore discuss components of the machine learning workflow that may be in need of interpretability, building on a range of ideas from human-centred explainability, with a particular focus on explainees, contrastive statements and explanatory processes. %
Our discussion reconciles and complements current research to help better navigate open questions -- rather than attempting to address any individual issue -- thus laying a solid foundation for a grounded discussion and future progress of explainable artificial intelligence and interpretable machine learning. %
We conclude with a summary of our findings, revisiting the human-centred explanatory process needed to achieve the desired level of algorithmic transparency and understanding in explainees.%
\end{abstract}

\keywords{%
Defining Explainability; %
Interpreting and Understanding Explanations; %
Human-Centred Perspective; %
Machine Learning Interpretability; %
Artificial Intelligence Explainability.%
}%

\maketitle

\begin{myyellowbox}
\textbf{Highlights}%
\begin{itemize}[topsep=0pt,label=\faLightbulbO,leftmargin=0cm,itemindent=.5cm,labelwidth=\itemindent,labelsep=0cm,align=left]%
    \item Explainability is an interactive process that engenders understanding, not knowledge.%
    \item Understanding is derived via reasoning applied to transparent insights into models.  %
    \item Transparency of insights and understanding are gauged on a continuous spectrum.      %
    \item Insights are interpreted under background knowledge mediated via operational context.%
\end{itemize}
\end{myyellowbox}
\vspace{1em}%
\begin{mygreenbox}
\textbf{Context \& Significance}\quad%
Automated decision-making systems based on artificial intelligence and machine learning algorithms are becoming ubiquitous across many domains of everyday life. %
Increasingly often, they are in a position to impact human affairs but their operation remains mostly unchecked, raising fairness, accountability and transparency concerns. %
Focusing on explainability, the past decade has offered rapid advances to the benefit of society, yet this research space still lacks a universal foundation. %
This work synthesises interdisciplinary findings to date and offers a clear path forwards.%
\end{mygreenbox}
\vspace{1em}%
\begin{mybluebox}
\textbf{Bigger Picture}\quad%
This work organises contributions to explainable artificial intelligence and interpretable machine learning from across sociology, philosophy and computer science. %
Such a comprehensive and interdisciplinary view enables us to identify common themes and establish a unified, scientific foundation for future research by defining explainability of predictive systems through its ingredients and roles, with a strong focus on humans and operational contexts. %
Our meta-analysis provides a high-level introduction to key concepts and a methodological framework to navigate them, thus complementing topic-oriented surveys. %
The contributions offer a focal point to guide development and evaluation of such techniques centred on human understanding and not just transfer of knowledge. %
This perspective should help to advance the field beyond an academic exercise and enable creation of truly explainable systems, accelerating proliferation of comprehensible models in scientific and social applications.%
\end{mybluebox}

\section{Great Expectations -- Explainable Machines\label{ch:intro-sec:intro}}%

Transparency, interpretability and explainability promote understanding and confidence. %
As a society, we strive for transparent governance and justified actions that can be scrutinised and contested. %
Such a strong foundation provides a principled mechanism for reasoning about fairness and accountability, which we have come to expect in many areas. %
Artificial Intelligence (AI) systems, however, are not always held to the same standards. %
This becomes problematic when data-driven algorithms power applications that either implicitly or explicitly affect people's lives, for example in banking, justice, job screenings or school admissions~\cite{o2016weapons}. %
In such cases, creating explainable predictive models or retrofitting transparency into pre-existing systems is usually expected by the affected individuals, or simply enforced by law. %
A number of techniques and algorithms are being proposed to this end; however, as a relatively young research area, there is no consensus within the AI discipline on a suite of technology to address these challenges.%

Building intelligible AI systems is oftentimes problematic, especially given their varied, and sometimes ambiguous, audience~\cite{kirsch2017explain,preece2018stakeholders}, purpose~\cite{hall2019systematic} and application domain~\cite{bhatt2020explainable}. %
While intelligent systems are frequently deemed (unconditionally) opaque, it is not a definitive property and it largely depends on all of the aforementioned factors, some of which fall beyond the consideration of a standard AI development lifecycle. %
Without clearly defined explainability desiderata~\cite{sokol2020explainability} addressing such diverse aspects can be challenging, in contrast to designing AI systems purely based on their predictive performance, which is often treated as a quality proxy that can be universally measured, reported and compared. %
In view of this disparity, many engineers (incorrectly) consider these two objectives as inherently incompatible~\cite{rudin2019stop}, thus choosing to pursue high predictive performance at the expense of opaqueness, which may be incentivised by business opportunities.%

While high predictive power of an AI system might make it useful, its explainability %
determines its acceptability. %
The pervasiveness of automated decision-making in our everyday life, some of it bearing social consequences, requires striking a balance between the two that is appropriate for what is at stake; for example, approaching differently a car autopilot and an automated food recommendation algorithm. %
Another domain that could benefit from powerful and explainable AI is (scientific) discovery~\cite{roscher2020explainable} -- intelligent systems may achieve super-human performance, e.g., AlphaGo~\cite{silver2017mastering}, however a lack of transparency renders their mastery and ingenuity largely unattainable. %
Such observations have prompted the Defense Advanced Research Projects Agency (DARPA) to announce the eXplainable AI (XAI) programme~\cite{gunning2016broad,gunning2017explainable} that promotes building a suite of techniques to %
(i) create more explainable models while preserving their high predictive performance and %
(ii) enable humans to understand, trust and effectively manage intelligent systems.%

To address these challenges, AI explainability and Machine Learning (ML) interpretability solutions are developed at breakneck speed, giving a perception of a chaotic field that may seem difficult to navigate at times. %
Despite these considerable efforts, a universally agreed terminology and evaluation criteria are still elusive, with many methods introduced to solve a commonly acknowledged but under-specified problem, and their success judged based on ad-hoc measures. %
In this paper we take a step back and re-evaluate the foundations of this research to organise and reinforce its most prominent findings that are essential for advancing this field, with the aim of providing a well-grounded platform for tackling DARPA's XAI mission. %
Our work thus reconciles and complements the already vast body of technical contributions, philosophical and social treatises, as well as literature surveys by bringing to light the interdependence of interdisciplinary and multifaceted concepts upon which explainable AI and interpretable ML are built. %
Our discussion manoeuvres through this incoherent landscape by connecting numerous open questions and challenges -- rather than attempting to solve any individual issue -- which is achieved through a comprehensive review of published work that acknowledges any difficulties or disagreements pertaining to these research topics.%

In particular, we review the notions of a \emph{black box} and \emph{opaqueness} in the context of artificial intelligence, and formalise \emph{explainability} -- the preferred term in our treatment of the subject (Section~\ref{sec:bbnes}). %
We discuss the meaning and purpose of explanations as well as identify theoretical and practical prerequisites for lifting unintelligibility of predictive systems, based on which we define explainability as a socially-grounded technology providing insights that lead to \emph{understanding}, which both conceptualises such techniques and fixes their evaluation criterion. %
Furthermore, we show how transparency, and other terms often used to denote similar ideas, can be differentiated from explainability -- both overcome opaqueness, but only the latter leads to understanding -- which we illustrate with decision trees of different sizes. %
While the premise of our definition is clear, understanding largely depends upon the explanation recipients, who come with a diverse range of background knowledge, experience, mental models and expectations~\cite{gregor1999explanations}. %
Therefore, in addition to technical requirements, explainability tools should also embody various social traits as their output is predominantly aimed at humans. %
We discuss these aspects of AI and ML explainers in Section~\ref{ch:intro-sec:humans}, which considers the role of an \emph{explanation audience} and people's preference for \emph{contrastive statements} -- XAI insights inspired by explainability research in the social sciences~\cite{miller2019explanation}. %
We also examine the social and bi-directional \emph{explanatory process} underlying conversational explanations between humans, highlighting the desire of explainees (i.e., the audience of explanations) to customise, personalise and contest various aspects of explanatory artefacts provided for opaque systems within a congruent interaction~\cite{sokol2018glass,sokol2020one}. %
We then connect these desiderata to validation and evaluation approaches for explainable AI and Interpretable ML (IML) techniques, arguing for studies that concentrate on assessing the understanding gained by the target audience in favour of other metrics.%

Next, in Section~\ref{sec:tradeoff}, we take a closer look at explainability by design (ante-hoc) and techniques devised to remove opaqueness from pre-existing black boxes (post-hoc and, often, model-agnostic), focusing on the latter type given that such methods are universally applicable to a wide variety models, which increases their potential reach and impact. %
While these explainers are appealing, their modus operandi can be an unintended cause of low-fidelity explanations that lack truthfulness with respect to the underlying black box~\cite{rudin2019stop}. %
Furthermore, their flexibility means that, from a technical perspective, they can be applied to any predictive model, however they may not necessarily be equally well suited to the intricacies of each and every one of them. %
Creating a faithful post-hoc explainer requires navigating multiple trade-offs reflected in choosing specific components of otherwise highly-modular explainability framework and parameterising these building blocks based on the specific use case~\cite{sokol2019blimey,sokol2020towards,sokol2020limetree,sokol2020tut}. %
These observations %
prompt us to revisit the disputed \emph{transparency--predictive power trade-off} and assess \emph{benefits} of interpretability that go beyond understanding of predictive algorithms and their decisions, such as their \emph{fairness} and \emph{accountability}.%

We continue our investigation in Section~\ref{sec:elephant} by assessing %
\emph{explainability needs for various parts of predictive systems} -- data, models and predictions -- as well as multiplicity and diversity of these, sometimes incompatible, insights. %
To this end, we use an \emph{explainability taxonomy} derived from the Explainability Fact Sheets~\cite{sokol2020explainability} to reason about such systems within a well-defined framework that considers both their social and technical requirements. %
Notably, it covers human aspects of explanations, thus giving us a platform to examine the audience (explainees), explanation complexity and fidelity, as well as the interaction mode, among many others. %
This discussion leads us to a high-level overview of landmark XAI and IML literature that highlights the interplay between various (often interdisciplinary and multifaceted) concepts popular in the field, thus painting a coherent perspective. %
Section~\ref{sec:summary} then summarises our main observations and contributions:%
\begin{itemize}
    \item we formally define explainability and catalogue other relevant nomenclature;%
    \item we establish a spectrum of opaqueness determined by the desired level of transparency and interpretability;%
    \item we identify important gaps in human-centred explainability from the perspective of current technology;%
    \item we dispute universality of post-hoc explainers given their complexity and high degree of modularity; and%
    \item we address explanation multiplicity through explanatory protocols for data, models and predictions.%
\end{itemize}
These insights pave the way for the development of more intelligible and robust machine learning and artificial intelligence explainers.%

\section{Defining Black-Box and Explainable Artificial Intelligence\label{sec:bbnes}}%

To avoid a common criticism of explainability research, we begin by discussing the concept of interpretability. %
To this end, we identify causes of opaqueness when dealing with intelligent systems and assess prerequisites for their understanding. %
In this setting we observe \textbf{shades of black-boxiness}: an interpretability spectrum determined by the extent of understanding exhibited by explainees, which, among others, is conditioned upon their mental capacity and background knowledge. %
We link this finding with various notions used in XAI and IML literature, a connection that helps us to fix the nomenclature and \textbf{define explainability} (our preferred term).%

The term \textbf{black box} can be used to describe a system whose internal workings are opaque to the observer -- its operation may only be traced by analysing its inputs and outputs~\cite{beizer1995black,bunge1963general}. %
Similarly, in computer science (including AI and ML) a black box is a (data-driven) algorithm that can be understood as an automated process that we cannot reason about beyond observing its behaviour. %
For AI in particular, \citet{rudin2019stop} points out two main sources of opaqueness: %
(i) a \emph{proprietary} system, which may be transparent to its creators, but operates as a black box; and %
(ii) a system that is too \emph{complex} to be comprehend by \emph{any human}. %
While the latter case concerns entities that are universally opaque for the \emph{entire population}, we argue that -- in contrast to a binary classification~\cite{dawkins2011tyranny} -- this definition of black boxes essentially establishes a (continuous) \emph{spectrum of understanding}. %
Notably, different levels and degrees of transparency and understandability have previously been pointed out and discussed in relation to individual elements of predictive systems, explainees' background knowledge and complexity of information conveyed by explanations, however these observations are often limited to multiple, hand-crafted discrete categories~\cite{marr1982vision,lipton2016mythos,arrieta2020explainable,roscher2020explainable,kim2021multi}.%

For example, a seminal inquiry into opaqueness of visual perception systems by \citet{marr1982vision} suggested three different levels at which information processing devices can be understood. %
The top tier is \emph{computational theory}, which concerns abstract specification of the problem at hand and the overall goal. %
It is followed by \emph{representation and algorithm}, which deals with implementation details and selection of an appropriate representation. %
The final level is \emph{hardware implementation}, which simply establishes physical realisation of the explained problem. %
To illustrate his framework, \citet{marr1982vision} argued that understanding why birds fly cannot be achieved by only studying their feathers: ``%
In order to understand bird flight, we have to understand aerodynamics; only then do the structure of feathers and the different shapes of birds' wings make sense.%
'' %
Nonetheless, he points out that these three tiers are only loosely related and some phenomena may be explained at only one or two of them, therefore it is important to identify which of these levels need to be covered in each individual case to arrive at understanding.%

\citeauthor{lipton2016mythos}'s categorisation of transparency~\cite{lipton2016mythos} -- which he defines as the ability of a human to comprehend the (ante-hoc) mechanism employed by a predictive model -- may be roughly seen as a modern interpretation of \citeauthor{marr1982vision}'s levels of understanding~\cite{marr1982vision}. %
The first of \citeauthor{lipton2016mythos}'s dimensions is \emph{decomposability}, which entails appreciation of individual components that constitute a predictive system, namely: input, parameterisation and computation; %
it can be compared to \citeauthor{marr1982vision}'s \emph{computational theory}. %
Next, \emph{algorithmic transparency} involves understanding the modelling process embodied by a predictive algorithm, which relates to \citeauthor{marr1982vision}'s \emph{representation and algorithm}. %
Finally, \emph{simulatability} enables humans to simulate a decisive process in vivo at the level of the entire model, capturing a concept similar to \citeauthor{marr1982vision}'s \emph{hardware implementation}. %
These three levels of \citeauthor{lipton2016mythos}'s notion of transparency span diverse processes fundamental to predictive modelling, and their understanding can offer a comprehensive, albeit predominantly technical, view of such systems.%

While not universally recognised, knowledge, perception and comprehension of a phenomenon undeniably depend upon the observer's cognitive capabilities and mental model, the latter of which is an internal representation of this phenomenon built on real-world experiences~\cite{kulesza2013too}. %
For example, \citet{kulesza2013too} outline a \emph{fidelity}-based understanding spectrum spanning two dimensions:%
\begin{description}[labelindent=2\parindent]%
    \item[completeness] captures how truthful the understanding is overall (\emph{generality}); and%
    \item[soundness] determines how accurate the understanding is for a particular phenomenon (\emph{specificity}).%
\end{description}
Therefore, a \emph{complete} understanding of an event from a certain domain is equivalently applicable to other, possibly unrelated, events from the same domain; for example, gravity in relation to a pencil falling from a desk. %
A \emph{sound} understanding, on the other hand, accurately describes an event without (over-)simplifications, which may result in misconceptions; for example, leaving a pencil on a slanted surface results in it falling to the ground. %
Striking the right balance between the two depends upon the observer and may be challenging to achieve: completeness without soundness is likely to be too broad, hence uninformative; and the opposite can be too specific to the same effect.%
\footnote{%
Note that comparable distinctions can be found across literature. %
For example, a differentiation between \emph{functional} and \emph{mechanistic} \emph{understanding}, where the former concerns ``functions, goals, and purpose[s]'' and the latter relies on ``parts, processes, and proximate causal mechanisms''~\cite{lombrozo2014explanation,lombrozo2019mechanistic}. %
A similar categorisation is also pertinent to \emph{knowledge}, which can either be \emph{declarative} or \emph{procedural}~\cite{gregor1999explanations} -- the former allows to recall facts (i.e., ``knowing that'') and the latter translates to skills that enable performing a cognitive task (i.e., ``knowing how'').%
}%

Within this space, \citet{kulesza2013too} identify two particularly appealing types of a mental model%
:%
\begin{description}[labelindent=2\parindent,leftmargin=4\parindent]
    \item[functional] which is enough to operationalise a concept but does not necessarily entail the understanding of its underlying mechanism (akin to The Chinese Room Argument~\cite{searle1980minds,penrose1989emperor}); and%
    \item[structural] which warrants a detailed understanding of how and why a concept operates.%
\end{description}
For example, a functional understanding of a switch and a light bulb circuit can be captured by the dependency between flipping the switch and the bulb lighting up. %
A structural understanding of the same phenomenon, on the other hand, may focus on the underlying physical processes, e.g., closing an electrical circuit allows electrons to ``move'', which heats up the bulb's filament, thus emitting light (note simplifications employed by this explanation). %
The former understanding is confined to operating a light switch, while the latter can be generalised to many other electrical circuits. %
Each one is aimed at a different audience and their complexity should be fine-tuned for the intended purpose as explanations misdirected towards an inappropriate audience may be incomprehensible. %
These observations lead us to argue that such a spectrum of understanding in human explainability can constitute a yardstick for determining explanatory qualities of predictive algorithms -- a link that has mostly been neglected in the literature, but which can help us to explicitly define popular XAI and IML terminology.%

A considerable amount of research into explainable AI and interpretable ML published in recent years appears to suggest that it is a freshly established field; however, in reality it is more of a renaissance~\cite{gregor1999explanations}. %
While work in this area indeed picked up the pace in the past decade, interest in creating transparent and explainable, data-driven algorithms dates back at least to the 1990s~\cite{rudin2019stop}, and further back to the 1970s if expert systems are taken into account~\cite{leondes2001expert,gregor1999explanations}. %
With such a rich history and the increased publication velocity attributed to the more recent re-establishment of the field, one may think that this research area has clearly defined objectives and a widely shared and adopted \textbf{terminology}. %
However, with an abundance of keywords that are often used interchangeably in the literature -- without precisely defining their meaning -- this is not yet the case. %
The most common terms include, but are not limited to:%
\begin{tasks}[label=\labelitemi,item-indent=3\parindent,label-offset=0pt,after-item-skip={\dimexpr\itemsep+\parsep}](4)%
    \task explainability,
    \task observability,
    \task transparency,
    \task explicability,
    \task intelligibility,
    \task comprehensibility,
    \task understandability,
    \task interpretability,
    \task simulatability,
    \task explicitness,
    \task justification,
    \task rationalisation,
    \task sensemaking,%
    \task insight,
    \task evidence,
    \task reason, and
    \task cause.
\end{tasks}
Other keywords -- such as function (of), purpose (of), excuse, consequence, effect, implication and meaning -- can also be found in non-technical explainability research~\cite[page 32]{achinstein1983nature}. %
Additionally, \emph{explanandum} or \emph{explicandum} often appear in XAI and IML literature, however these terms, which are borrowed from philosophy of science, denote the concept to be explained.%

While early XAI and IML research might have missed out on an opportunity to clearly define its goals and nomenclature, recent work has attempted to rectify this problem~\cite{%
biran2014justification,biran2017explanation,rudin2019stop,gilpin2018explaining,alvarez2019weight,mohseni2021multidisciplinary,chen2022machine,lipton2016mythos,vilone2020explainable,markus2021role,guidotti2018survey,langer2021we,rosenfeld2019explainability,bibal2016interpretability,papenmeier2022complicated,montavon2018methods,yao2021explanatory,furnkranz2020cognitive,murdoch2019definitions,roscher2020explainable,kim2021multi,adadi2018peeking,offert2017know,arrieta2020explainable,marcinkevivcs2020interpretability,kaur2022sensible,stepin2021survey,palacio2021xai%
}. %
Within this spaces, some authors simply list the relevant terms without assigning any meaning to them~\cite{marcinkevivcs2020interpretability} while others cite dictionary definitions~\cite{roscher2020explainable,arrieta2020explainable,palacio2021xai} -- e.g., to interpret is ``to explain [\ldots] the meaning of'' or ``present in understandable terms'' according to Merriam-Webster -- or suggest that many such keywords are synonymous~\cite{bibal2016interpretability}. %
It is also common to find circular or tautological definitions, which use one term from the list shown above to specify another~\cite{adadi2018peeking}; %
for example, ``something is explainable when we can interpret it'', ``interpretability is making sense of ML models'', ``interpretable systems are explainable if their operations can be understood by humans'' or ``intelligibility is the possibility to comprehended something''. %
Hierarchical and ontological definitions of these terms also appear in the literature~\cite{rosenfeld2019explainability,bibal2016interpretability,stepin2021survey}, often creating a web of connections that is difficult to parse, follow and apply. %
Another counterproductive approach to defining these concepts assumes that their meaning is inherently intuitive or can only be captured by tacit knowledge -- viewpoints that can be summarised by ``I know it when I see it'' phrase~\cite{offert2017know}.%

A different route to specifying these terms is binding keywords to particular components of a predictive pipeline or associating them with the technical and social properties of such a system~\cite{markus2021role,lipton2016mythos,roscher2020explainable}; however, the former is just a labelling strategy and the latter assumes that we can achieve explainability by simply fulfilling a set of requirements. %
A fictitious scenario under this purview could determine that data are understandable, models are transparent and predictions are explainable; %
the overall interpretability of predictive pipelines, on the other hand, is determined by the fidelity, brevity and relevance of the insights produced by the designated method. %
More specifically, transparency is oftentimes associated with ante-hoc methods (the internal mechanisms of a predictive model) and interpretability with post-hoc approaches (the observable behaviour a system)~\cite{lipton2016mythos}; %
alternatively, interpretability is designated for models and explainability for their outputs (more precisely, reasons behind predictions)~\cite{kim2021multi}. %
Similarly, simulatability may be linked to an entire model, decomposability to its individual components and algorithmic transparency to the underlying training algorithm~\cite{lipton2016mythos}. %
Interpretability has also been defined as a domain-specific notion that imposes ``a set of application-specific constraints on the model'', thus making this concept only applicable to predictive models that can provide their own explanations (i.e., ante-hoc interpretability) and should not, along with the term explainability, be used to refer to ``approximations to black box model predictions'' (i.e., post-hoc explainability)~\cite{rudin2019stop}. %
In this particular view, therefore, a predictive model is interpretable if it ``obeys structural knowledge of the domain, such as monotonicity, causality, structural (generative) constraints, additivity or physical constraints that come from domain knowledge''.%

Alternatively, interpretability could be used as an umbrella term and refer to data, models and post-hoc approaches as long as it facilitates extracting helpful information from these components or produces insights into them~\cite{murdoch2019definitions}. %
This designation, however, may not be universal and, to complicate matters even more, certain terms can be used with respect to multiple elements of a predictive system, causing confusion. %
For example, transparency may relate to predictive models, interpretability to input data and models, and explainability to data, models and human recipients~\cite{roscher2020explainable}. %
While transparency may be uniquely linked to predictive models, it can still carry multiple meanings and apply to operations of a model as well as its design, components and the underlying algorithmic process~\cite{roscher2020explainable,lipton2016mythos}. %
Desiderata-based definitions appearing in the literature, on the other hand, specify explainability through a mixture of properties such as interpretability (determined by clarity \& parsimony) and fidelity (consisting of completeness \& soundness)~\cite{markus2021role,rosenfeld2019explainability}; nonetheless, note that achieving both does not necessarily guarantee better comprehension of the explained system. %
In like manner, \citet{alvarez2019weight} used the \emph{weight of evidence} idea from information theory to mathematically define AI and ML explainability and outline a precise list of its desiderata. %
While appealing, the complexities of the real world make their conceptualisation difficult to apply at large. %
\citet{murdoch2019definitions} followed a similar route and captured interpretation, understanding and evaluation of data-driven systems in the predictive--descriptive--relevant (PDR) framework, which spans: predictive performance (accuracy) of a model; descriptive capabilities of its (post-hoc) explainer quantified via fidelity; and relevancy of the resulting explanatory insights determined by their ability to provide ``insight[s] for a particular audience into a chosen problem''.%

Within this chaotic landscape some researchers propose flexible definitions that are inspired by interdisciplinary work and can accommodate a variety of contexts while maintaining a precise and coherent meaning. %
\citet{gilpin2018explaining} offered definitions of ``explanation'', ``interpretability'' and ``explainability'' drawing from a broad body of literature in an effort to standardise XAI and IML findings. %
While their notions appear somewhat vague -- explanations should answer ``Why?'' and ``Why-should?'' questions until such questions can no longer be asked -- they argue for making explanations \emph{interpretable} and \emph{complete}, striving towards human \emph{understanding} that depends on the explainee's cognition, knowledge and biases. %
Similarly, \citet{biran2014justification} were concerned with \emph{explanations}, which they characterised as ``giving a reason for a prediction'' and answering ``how a system arrives at its prediction''. %
They also defined \emph{justifications} as ``putting an explanation in a context'' and conveying ``why we should believe that the prediction is correct'', which, they note, do not necessarily have to correspond to how the predictive system actually works. %
Notably, many of these observations reappear across diverse publications, with the shared theme indicating that explanations should always answer an implicit or explicit ``Why?'' question~\cite{koura1988approach}, in addition to addressing ``What?'' and ``How?''~\cite{miller2019explanation,palacio2021xai}. %
In a later piece of work, \citet{biran2017explanation} defined explainability as ``the degree to which an observer can understand the cause of a decision'' (also adopted by \citet{miller2019explanation}), thus making it much more explainee-centred. %
While many authors use the term \emph{cause} rather loosely in XAI and IML research, we argue against such practice -- it is important to reserve it exclusively for insights extracted from \emph{causal} models~\cite{pearl2018book}.%

More recently, based on an extensive review of literature in computer science and related disciplines, \citet{mohseni2021multidisciplinary} provided a collection of definitions for the most common terms in explainable AI and interpretable ML, nonetheless the underlying rationale is predominantly qualitative making them difficult to operationalise. %
Similarly, \citet{arrieta2020explainable} differentiated between the following terms: understandability/intelligibility, comprehensibility, interpretability/transparency and explainability. %
Specifically, they defined interpretability or transparency as a \emph{passive} characteristic of a model that allows humans to make sense of it on different levels -- e.g., its internal mechanisms and derivation of predictions -- therefore relating it to the cognitive skills, capacities and limitations of individual explainees. %
Explainability, on the other hand, was described as an \emph{active} characteristic of a model that is achieved through actions and procedures employed (by the model) to clarify its functioning for a certain audience. %
\citet{montavon2018methods} also offered definitions of these two terms -- interpretability and explainability -- from a perspective of functional understanding~\cite{lombrozo2014explanation,lombrozo2019mechanistic}. %
They characterised interpretability as a mapping of an abstract concept, e.g., a predicted class, onto a domain that can be comprehended by a human; %
explainability, in their view, is responsible for providing a collection of factors -- expressed in an interpretable domain -- that contribute to an automated decision of a particular instance. %
In summary, their goal is to study and understand how inputs are mapped to outputs, possibly via a human-comprehensible representation of relevant concepts.%

Each definition conveys a more or less precise meaning that can be used to label relevant techniques, however they do not necessarily clarify and help to navigate the complex landscape of IML and XAI research. %
To organise this space, we categorise the underlying terminology based on three criteria:%
\begin{itemize}
    \item \emph{properties} of systems;%
    \item \emph{functions} and \emph{roles} that they serve; and%
    \item \emph{actions} required to process, assimilate and internalise information elicited by them.%
\end{itemize}
The core concept around which we build our nomenclature is \textbf{explainability}; we define it as \textbf{insights that lead to understanding} (the \textbf{role} of an explanation) -- a popular and widely accepted rationale in the social sciences~\cite{koehler1991explanation,baumeister1994self,lombrozo2006structure,paez2019pragmatic,woodward1979scientific,achinstein1983nature}. %
While it may seem abstract, understanding can be assessed with questioning dialogues~\cite{walton2007dialogical,walton2011dialogue,walton2016dialogue,arioua2015formalizing,madumal2019grounded} -- e.g., a machine interrogating the explainees to verify their understanding of the phenomenon being explained at the desired level of detail -- which are the opposite of explanatory dialogues. %
Such a process reflects how understanding is tested in education, where the quality of tuition as well as knowledge and skills of pupils are evaluated through standardised tests and exams (albeit not without criticism~\cite{mead2015teacher}). %
Furthermore, encouraging people to explain a phenomenon helps them to realise the extent of their ignorance and confront the complexity of the problem, which are important factors in exposing The Illusion of Explanatory Depth~\cite{rozenblit2002misunderstood} -- a belief that one understands more than one actually does.%

This notion of explainability and the three building blocks of XAI and IML terminology allow us to precisely define the other popular terms. %
Therefore,%
\begin{tasks}[label=\labelitemi,item-indent=3\parindent,label-offset=0pt,after-item-skip={\dimexpr\itemsep+\parsep}](3)%
    \task \emph{observability},%
    \task \emph{transparency},%
    \task \emph{explicability},%
    \task \emph{intelligibility},%
    \task \emph{comprehensibility},%
    \task \emph{understandability},%
    \task \emph{interpretability},%
    \task \emph{simulatability}, and%
    \task \emph{explicitness}%
\end{tasks}
are \textbf{properties} of an AI or ML system %
that enable it to %
directly (ante-hoc) or indirectly (post-hoc) %
convey information of varied complexity, the \emph{understanding} of which depends upon the cognitive capabilities and (domain) expertise of the explainee. %
For example, observing an object falling from a table is a transparent phenomenon per se, but the level of its understanding, if any, is based upon the depth of the observer's physical knowledge. %
Such characteristics provide%
\begin{tasks}[label=\labelitemi,item-indent=3\parindent,label-offset=0pt,after-item-skip={\dimexpr\itemsep+\parsep}](3)%
    \task \emph{justification},%
    \task \emph{rationalisation},%
    \task \emph{insight},%
    \task \emph{evidence}, and%
    \task \emph{reason}%
\end{tasks}
(\textbf{roles}) that can be used to%
\begin{tasks}[label=\labelitemi,item-indent=3\parindent,label-offset=0pt,after-item-skip={\dimexpr\itemsep+\parsep}](3)%
    \task \emph{reason} about,%
    \task \emph{make sense} of,%
    \task \emph{rationalise},%
    \task \emph{justify},%
    \task \emph{interpret}, or%
    \task \emph{comprehend}%
\end{tasks}
(note that here these are used as verbs) behaviour of a -- black-box or glass-box -- predictive system, all of which are \textbf{actions} that under the right circumstances lead to \emph{understanding}. %
While \emph{simulatability} is also based upon observing a transparent process and replicating it, such an \textbf{action} does not necessarily imply understanding of the underlying phenomenon -- recall the difference between declarative and procedural knowledge~\cite{gregor1999explanations}, structural and functional mental models~\cite{kulesza2013too}, functional and mechanistic understanding~\cite{lombrozo2014explanation,lombrozo2019mechanistic} and The Chinese Room Argument~\cite{searle1980minds} discussed earlier. %
Lastly, a \emph{cause} has a similar meaning to a \emph{reason}, but the first one is derived from a causal model, whereas the latter is based purely on observations of the behaviour of a (black-box) model.%

Such a setting makes a welcome connection between the XAI and IML terminology synthesised by the equation%
\[%
\texttt{Explainability} \; = \; %
\underbrace{%
\texttt{Reasoning} \left( \texttt{Transparency} \; | \; \texttt{Background Knowledge} \right)%
}_{\textit{understanding}}%
\text{,}%
\]%
which defines \texttt{Explainability} as the \textbf{process} of deriving \emph{understanding} -- i.e., extracting meaning -- through \texttt{Rea\-son\-ing} applied to \texttt{Transparent} insights distilled from a data-driven predictive system that are adjusted to the explainee's \texttt{Background Knowledge}. %
In this process, the \texttt{Reasoning} can either be done by the explainer or the explainee, and there is an implicit assumption that the explainee's \texttt{Background Knowledge} aligns with the \texttt{Transparent} representation of the predictive model. %
If the latter does not hold, mitigation techniques such as employing an \emph{interpretable representation} can be used to communicate concepts that are otherwise incomprehensible~\cite{ribeiro2016why,sokol2019blimey,sokol2020towards}. %
\texttt{Reasoning} also comes in many different shapes and sizes depending on the underlying system (\texttt{Transparency}) as well as the explainer and the explainee (\texttt{Background Knowledge}); for example, %
logical reasoning with facts, %
causal reasoning over a causal graph, %
case-based reasoning with a fixed similarity metric, and %
artificial neuron activation analysis for a \emph{shallow} neural network.%

Therefore, linear models are transparent given a reasonable number of features; additionally, with the right ML and domain background knowledge -- requirement of normalised features, effect of feature correlation and the meaning of coefficients -- the explainee can reason about their properties, leading to an explanation based on understanding. %
Similarly, a visualisation of a \emph{shallow} decision tree can be considered both transparent and explainable assuming that the explainee understands how to navigate its structure (ML background knowledge) and the features are meaningful (domain background knowledge); again, it is up to the explainee to reason about these insights. %
When the size of a tree increases, however, its visualisation loses the explanatory power because many explainees become unable to process and reason about its structure. %
Restoring the explainability of a deep tree requires delegating the reasoning process to an algorithm that can digest its structure and output sought after insights in a concise representation. %
For example, when explaining a prediction, the tree structure can be traversed to identify a similar instance with a different prediction, e.g., as encoded by two neighbouring leaves with a shared parent, thus demystifying the automated decision~\cite{sokol2019desiderata,sokol2021towards}.%

While understanding and applying the newly acquired knowledge to unseen tasks are recurring themes in XAI and IML literature~\cite{doshi2017towards,bibal2016interpretability,kim2021multi,biran2017explanation,furnkranz2020cognitive,chen2022machine,rosenfeld2019explainability} -- with a few notable exceptions~\cite{palacio2021xai,yao2021explanatory,langer2021we} -- they rarely ever play the central role. %
For example, \citet{palacio2021xai} define an explanation as ``the process of describing one or more facts, such that it facilitates the understanding of aspects related to said facts (by a human consumer)''. %
Similarly, \citet{yao2021explanatory} ``highlight[s] one important feature of explanations: they elicit understanding (in humans)''; %
\citeauthor{yao2021explanatory} proceeds to suggest that the three levels of analysis proposed by \citet{marr1982vision}, and discussed earlier, should be extended with a \emph{social} level to reflect that AI models, especially the ones of concern to XAI and IML researchers, do not operate in isolation from humans.%

Furthermore, some researchers appear to converge towards a similar definition to ours. %
\citet{papenmeier2022complicated} characterise transparency as ``the extent to which information is exposed to a system's inner workings'' and interpretability as ``the extent to which transparency is meaningful to a human'', both of which lead them to formalise explanations as ``the mechanisms by which a system communicates information about its inner workings (transparency) to the user''. %
In like manner, \citet{roscher2020explainable} posit that ``the scientist is using the data, the transparency of the method, and its interpretation to explain the output results (or the data) using domain knowledge and thereby [\ldots] obtain[s] a scientific outcome'', which process should lead to understanding by presenting properties of ML models in humans-comprehensible terms; %
they further suggest that ``explainability usually cannot be achieved purely algorithmically'', which resonates with the role of human \emph{reasoning} in our definition. %
Similarly, \citet{langer2021we} identify a ``(given) context'' as the element moderating ``explanatory information'' to facilitate ``stakeholders' understanding'' (of a subset of components present in a complex system). %
Additionally, while not explicitly stated, one interpretation of \citeauthor{rosenfeld2019explainability}'s notion of explanations~ \cite{rosenfeld2019explainability} is a collection of human-centred processes that allow explainees to understand a predictive model by presenting them with a suitable representation of its logic (a concept that they call \emph{explicitness}), however based on a graphical representation of their framework explainability is achieved through interpretability, only one component of which is transparency. %
Adjacent to XAI and IML, \citet{bohlender2019towards} discussed explanations in software systems, which are meant to ``resolve [a] lack of understanding'' pertaining to a particular aspect (explanandum) of a system for a specific audience and ``the processing of [an] explanation [\ldots] is what makes [an] agent [\ldots] understand the explanandum'' -- an action that may require the use of cognitive or computational resources and that may only be operationalised in specific contexts determined, for example, by the background knowledge of the explainees.%

Given the importance of \emph{understanding} in our definition, as well as explainability research outside of XAI and IML~\cite[pages 23, 42, 57]{achinstein1983nature}, it is crucial to review its acquisition and operationalisation together with how it is distinct from and more desirable than not only declarative but also procedural \emph{knowledge}. %
For example, consider justifications that can be seen to communicate why a decision is correct without necessarily providing the exact logic behind it~\cite{biran2017explanation}, therefore preventing the explainee from internalising and reapplying these insights to other scenarios. %
Similarly, knowledge can be acquired and recalled thus giving the impression of understanding but recitation in itself does not imply comprehension or operationalisation (i.e., an ability to apply it) -- an observation that follows from The Chinese Room Argument~\cite{searle1980minds}. %
For example, memorising a textbook or answers to a set of questions may suffice to pass an exam but falls short of effectively resolving related yet distinct problems~\cite{gregor1999explanations}. %
This is an important insight for evaluating explainability -- as defined in this paper -- since we first have to specify what it means to attain understanding before we can assess effectiveness of the explanatory process~\cite{paez2019pragmatic}. %
Furthermore, internalising knowledge into understanding and correctness thereof are conditioned on the reasoning capabilities and background knowledge of the explainee~\cite{achinstein1983nature,gregor1999explanations}. %
The detrimental effects of misalignments in this space were shown by \citet{bell2022its}, who reported that explainability mechanisms pertinent to inherently interpretable models may be confusing, especially so for a lay audience; %
for example, ante-hoc transparency of decision trees achieved though a visualisation of their structure misleads people (lacking technical expertise) into believing that the feature used by the root-node split is the most important attribute.%

\section{Humans and Explanations -- Two Sides of the Same Coin\label{ch:intro-sec:humans}}

Defining explainability as leading to understanding and our categorisation into \emph{properties}, \emph{functions} and \emph{actions} highlight an important aspect of this research topic: explanations do not operate in a vacuum, they are highly contextual and directed at some autonomous agent, either a human or machine, who is as important as the explainability algorithm itself. %
Notably, up until recently XAI and IML research has been undertaken mostly within the computer science realm~\cite{miller2017explainable}, thus bringing in various biases and implicit assumptions from this predominantly technical field. %
While some explainability research has found its way into other scientific disciplines, e.g., law~\cite{wachter2017counterfactual}, the majority gravitated around technical properties. %
This research agenda was disrupted by \citet{miller2017explainable}, who observed that the function of an explanation and its recipients are largely neglected -- a phenomenon which they dubbed ``inmates running the asylum'' -- leading to a substantial paradigm shift. %
\citeauthor{miller2019explanation}'s follow-on work~\cite{miller2019explanation} grounded this observation in (human) explainability research from the social sciences, where this topic has been studied for decades, thus providing invaluable insights that can benefit XAI and IML.%

\citeauthor{miller2019explanation}'s findings~\cite{miller2019explanation} have arguably reshaped the field, with a substantial proportion of the ensuing research acknowledging the \textbf{explainees} -- their autonomy, goals, expectations, intentions and interactions. %
While explainability of data-driven systems has various benefits, it is usually in focus when an AI agent fails, behaves anomalously or operates inconsistently with the explainee's beliefs, expectations or mental model, e.g., an unexpected ML prediction causing a disagreement; alternatively, an explanation may be requested to support learning or provide information needed to solve a problem or complete a task~\cite{gregor1999explanations}. %
In such cases, explainees' preferences, needs and goals should be addressed to maximise the effectiveness of an explanation, for example by appropriately adjusting its complexity~\cite{miller2019explanation,gregor1999explanations}. %
This step can be further improved by treating explainability as a process instead of one-off information offloading~\cite{miller2019explanation,gregor1999explanations}; by satisfying the explainees' natural desire to \textbf{interact} and communicate with the explainer within a predictable protocol, they are provided with an opportunity to seamlessly customise and personalise the explanation~\cite{sokol2020one}. %
Perhaps the most influential of \citeauthor{miller2019explanation}'s observations is the humans' preference for \textbf{contrastive} explanations given their prominence in everyday life. %
We discuss these three fundamental aspects of human-centred explainability in more detail below.%

Understanding can be an elusive objective when it comes to explaining intelligent systems since each unique \textbf{explanation audience} may expect to receive different insights, e.g., a medical diagnosis can be communicated in terms of test results or observable symptoms depending on whether it is directed towards medical staff or patients. %
While in our considerations we implicitly assume that the explanation recipient is a human, it may as well be another algorithm that further processes such insights, in which case other, more suitable, properties would be of interest. %
When taken into account, the \emph{purpose} of explainability and the explainee's goal also influence the explanation composition~\cite{gregor1999explanations}. %
For example, an explanation will look different when it helps to debug an ML model and when it justifies a negative outcome of a loan application; note that the target audience also differs, with the former aimed at ML engineers and the latter at lay people. %
A complementary view, based on a \emph{means-end account} of XAI~\cite{buchholz2022means}, argues to examine ``\emph{what} should be explained (topic), \emph{to whom} something should be explained (stakeholder), \emph{why} something should be explained (goal), and \emph{how} something should be explained (instrument)''. %
Addressing such desiderata by accounting for the explainee's cognitive capabilities and skill level, however, is challenging as it requires access to the explainee's background knowledge and mental model, which are vague and often undefined concepts that cannot be easily extracted and modelled.%

Nonetheless, just by considering the audience and purpose of an explanation, we can identify (and deliver) a collection of relevant properties. %
In certain cases, such as the aforementioned loan application, the \emph{actionability} of explanatory insights is crucial, e.g., suggesting that an individual would receive a loan had he or she been 10 years younger is futile. %
Multiplicity of apparently indistinguishable arguments can also decrease the perceived quality of an explanation when one is chosen at random without a user-centred heuristic in place, which, again, depends on the application domain and audience. %
For example, research suggests~\cite{miller2019explanation} that if one of multiple, otherwise equivalent, \emph{time}-ordered events has to be chosen as an explanation, the most recent one will best resonate with the explainee; additionally, prioritising explanations by their \emph{novelty} will keep the explainee engaged and attentive, and distinguishing between \emph{sufficient} and \emph{necessary} conditions for a given outcome can help to reduce cognitive load. %
While desirable, \emph{brevity} of an explanation can sometimes be at odds with its comprehensiveness and \emph{completeness} -- sacrificing the big picture (which in itself may be too convoluted to understand) for concise communication~\cite{kulesza2015principles}. %
Explanatory minimalism, nonetheless, bears the danger of oversimplification; however, when it %
is a strict requirement, explanation \emph{soundness} can be favoured to focus on factors pertinent to the explained instance and discard more general reasons that are largely irrelevant. %
Such an approach can introduce inaccuracies with respect to the overall data-driven system, but the explanations remain truthful for the individual instance. %
Striking the right balance between generality and specificity of an explanation -- as well as achieving all the other aforementioned desiderata -- is notoriously challenging and often requires tuning its soundness and completeness for the intended audience and application, which itself may be impractical when done manually, and prohibitively difficult through capturing the explainee's mental model.%

While posing problems for AI explainers, satisfying this wide range of diverse assumptions and expectations comes naturally to humans when they engage in an \textbf{explanatory process} among themselves. %
This is partly due to shared background knowledge, and is further amplified by interactive communication that allows to rapidly iterate over questions, exchange informations and refine answers to arrive at understanding. %
One explanation does not fit all and treating explainability as a bi-directional process provides a platform to appreciate uniqueness of each explainee through personalised explanations~\cite{sokol2020one} that enable transfer of knowledge and help to develop understanding~\cite{gregor1999explanations}. %
While these topics have received relatively little attention in the XAI and IML literature, we can draw design insights and inspirations from research on \emph{explanatory debugging} of predictive models~\cite{kulesza2015principles}. %
Therefore, an interactive explanatory process should be \emph{iterative}, enabling the explainee to learn, provide feedback and receive updated insights until reaching a satisfactory conclusion; %
the explainer ought to always \emph{honour user-provided feedback} by incorporating it into the explanation generation process, or clearly communicate a precise reason if that is impossible; %
the communication should be \emph{reversible} to %
allow the explainee to retract a requested change or annul a piece of feedback when it was provided by mistake, or to explore an interesting part of the predictive model through a speculative enquiry; and, finally, %
the whole process should be \emph{incremental} to easily attribute each piece of feedback to an explanation change, thereby showing up-to-date results regardless of how small the tweaks are.%

Even though dialogue is fundamental to human explainability, it is largely absent in XAI and IML techniques~\cite{sokol2020one}, which are often based on one-way communication, where the user receives a description of a data-driven system without an opportunity to request more details or contest it. %
A similar interaction in a form of the aforementioned questioning dialogues can also be used to judge the explainee's understanding of the explained concept, thus be a proxy for assessing effectiveness of the explainer. %
Notably, human dialogue tends to be verbal or written, both of which are based on the natural language. %
While ubiquitous, this form of communication is not equally effective in conveying all types of information, requiring humans to augment it with visual aids, which are especially helpful when the interaction serves explanatory purposes. %
The same strategy can be adopted in explainable AI and interpretable ML, where the explainer would switch between various explanatory artefacts -- such as (written and spoken) text, images, plots, mathematical formulation, numbers and tables -- depending on which one is best suited for the type of information being communicated in a given context~\cite{gregor1999explanations}. %
Mixing and matching them is also possible, e.g., a numerical table or a plot complemented with a caption, and may be beneficial as the whole can be greater than the sum of its parts, especially that certain explanation types may require a specific communication medium or explanatory artefact to be effective. %
Using visualisation, textualisation and (statistical) summarisation, however, does not guarantee a coherent relation, structure or story conveyed by these communication media alone, which could possibly be achieved by grounding them in a shared context through \emph{logical reasoning} or \emph{formal argumentation}~\cite{dung2009assumption}; %
additional inspiration can be found in \emph{sensemaking theory}, which was recently adapted to XAI and IML applications~\cite{kaur2022sensible}.%

\textbf{Contrastive explanations} -- more specifically, counterfactuals -- dominate the human explanatory process and are considered the gold standard in explainability of data-driven systems~\cite{miller2019explanation}. %
They juxtapose a hypothetical situation (foil) next to the factual account %
with the aim to emphasise the consequences of or ``would be'' change in the outcome. %
Counterfactuals statements can be characterised by their lineage: \emph{model-driven} explanations are represented by artificial data points (akin to centroids), whereas \emph{data-driven} explanations are instances recovered from a (training) data set (similar to medoids). %
Furthermore, the contrast can either be implicit -- i.e., ``Why class \(X\)?'' (hence not any other class) -- or explicit -- i.e., ``Why class \(X\) and not \(Y\)?'' %
Counterfactuals are appropriate for lay audiences and domain experts alike, can use concepts of varying difficulty and be expressed in different media such as text and images. %
They are parsimonious as the foil tends to be based on a single factor, but, if desired, can account for an arbitrary degree of feature covariance. %
They support interaction, customisation and personalisation, e.g., a foil built around a user-selected feature provided in an explanatory dialogue, which can be used to restrict their search space, possibly making them easier to retrieve. %
When deployed in a user-centred application, they can provide the explainees with appealing insights by building the foil only around actionable features. %
However, their effectiveness may be problematic when explaining a proprietary predictive system, e.g., built as a black box with the intention to protect a trade secret, since counterfactual explanations can leak sensitive information, thereby allowing the explainee to steal or game the underlying model. %
In an open world, they also suffer from vaguely defined or imprecise notions known as \emph{non-concepts}~\cite{offert2017know}, e.g., ``What is not-a-dog?''%

These idealised properties make counterfactual statements appealing, but some may get lost in practice, e.g., an imperfect implementation, resulting in subpar explanations. %
On the face of it, these explanatory artefacts resemble causal insights, but unless they are generated with a causal model~\cite{pearl2016causal}, they should not be treated as such and instead be interpreted as descriptors of a decision boundary used by a predictive system. %
If they are model-driven, as opposed to data-driven, they may not necessarily come from the data manifold, yielding (out-of-distribution) explanations that are neither feasible nor actionable in the real life, e.g., ``Had you been 200 years old, \ldots'' %
Even if they are consistent with the data distribution, the foil may still come from a sparse region, thus prescribing possible but improbable feature values~\cite{poyiadzi2020face}. %
Counterfactual explanations are often specific to a single data point, although humans are known to generalise such insights to unseen and possibly unrelated cases -- recall The Illusion of Explanatory Depth effect~\cite{rozenblit2002misunderstood} -- which may result in overconfidence.%

Fulfilling all of these desiderata can help in developing an explainability system that enables the explainees to (better) \emph{understand} an automated decision-making process, which, as we noted earlier, offers a yardstick by which success of such techniques can be quantified~\cite{kim2021multi,paez2019pragmatic,bohlender2019towards,langer2021we}. %
To be effective, however, the approach to assess, evaluate and measure understanding has to account for the aforementioned contextual parameters such as the application domain (and its sensitivity), function of the explanation, intended audience (and its background knowledge) as well as (technical) caveats of the XAI or IML algorithm~\cite{arrieta2020explainable,murdoch2019definitions}. %
Just like beauty, which is in the eye of the beholder, the (perceived) quality, helpfulness and success of an explanation are judged by the recipient~\cite{kim2021multi}. %
Therefore, producing explanations is a necessary but insufficient condition of engendering understanding, which additionally requires them to be relevant to the stakeholders, comprehensible by them and compatible with their desiderata~\cite{langer2021we,gregor1999explanations}, %
for example, by aligning the type of insights and their level of transparency towards the chosen use case and audience (in view of the anticipated background knowledge and skills). %
Specifically, consider the truthfulness of explanatory information, which may be reduced for a particular purpose without harming the audience's ability to derive understanding; %
the simplification used in depicting underground lines and stops is a case in point as it conveys the cues necessary to navigate such a transportation system despite foregoing an accurate representation of scale and distance~\cite{paez2019pragmatic}. %
Notably, a transparent representation that is universal and satisfies all the different needs and expectations of diverse explainee groups may not exist, just like an agreement between these individuals on the objective nature of understanding. %
Given this strong dependence on human perception, the effectiveness of explanations should be evaluated empirically~\cite{bohlender2019towards} to combat The Illusion of Explanatory Depth~\cite{rozenblit2002misunderstood}, and, in view of The Chinese Room Argument~\cite{searle1980minds}, the studies should go beyond assessing simple task completion to capture the difference between knowledge and understanding within a well-defined, real-life context expected in the deployment.%

\section{The Discord over Sacrificing Explainability for Predictive Power\label{sec:tradeoff}}%

Theoretical desiderata do not always align with the operationalisation and practicalities of XAI and IML algorithms, and the latter are what ends up affecting our lives. %
For example, explainability is an inherently social process that usually involves bi-directional communication, but most implementations -- even the ones using contrastive statements~\cite{wachter2017counterfactual,waa2018contrastive} -- output a single explanation that is optimised according to some predefined metric, not necessarily addressing concerns of an individual explainee~\cite{sokol2020one}. %
Similarly, while inherently transparent predictive models and ante-hoc explainers may be preferred~\cite{rudin2019stop}, such solutions are often model-dependent, labour-intensive and tend to be application-specific, which limits their scope as well as wider applicability and adoption. %
Instead, post-hoc and model-agnostic explainers dominate the field~\cite{ribeiro2016why,lundberg2017unified,ribeiro2018anchors,sokol2019blimey} since they are considered one-stop solutions -- a unified explainability experience without a cross-domain adaptation overhead. %
This silver bullet framework, however, comes at a cost: subpar fidelity that can result in misleading or outright incorrect explanations. %
While increasingly such considerations find their way into publications, they are often limited to acknowledging the method's shortcomings, stopping short of offering a %
\begin{wrapfigure}[12]{i}{.28\textwidth}%
    \centering
    \vspace{-.25\baselineskip}
    \includegraphics[width=.225\textwidth]{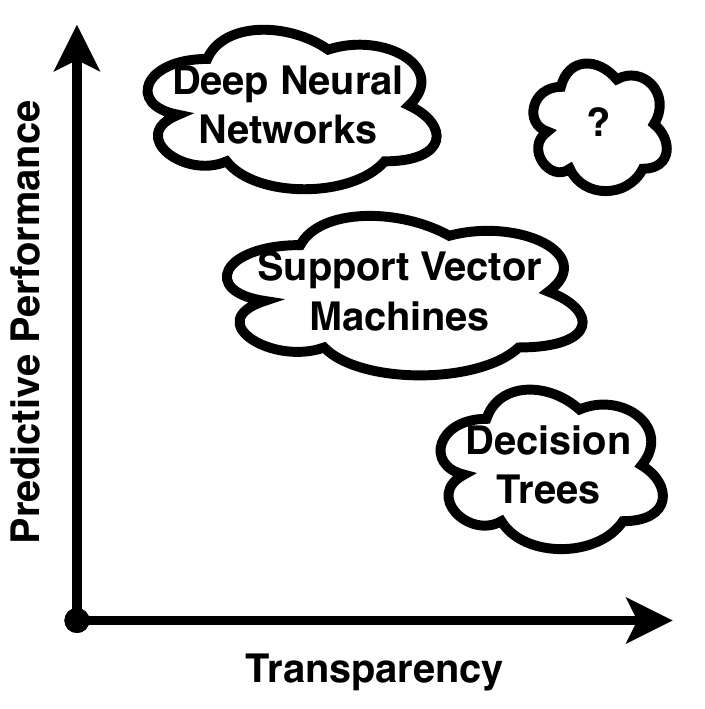}%
    \caption{%
        Fictitious depiction of the anecdotal trade-off between transparency and predictive power of AI systems~\cite{gunning2016broad}.%
        \label{fig:ch1:tradeoff}%
    }
\end{wrapfigure}
viable solution.%

A common belief motivating many methods published in the explainability literature is the perceived \emph{dichotomy} between transparency and predictive power of AI and ML systems. %
A popular example supporting this theory is the unprecedented effectiveness of deep neural networks on certain tasks, whose ever increasing complexity, e.g., the number of layers %
and hidden units, improves their performance at the expense of transparency. %
This trade-off has been reiterated in the DARPA XAI program's Broad Agency Announcement~\cite{gunning2016broad} and supported by an appealing graph reproduced in Figure~\ref{fig:ch1:tradeoff}. %
However, at the time of publication it has been a \emph{theory} based mostly on anecdotal evidence, with \citet{rudin2019stop} criticising plots like this given their lack of scale, transparency or precise performance metrics, and supporting data. %
Notably, \citeauthor{rudin2019stop} argues that investing more effort into feature engineering and data modelling can help to build inherently explainable AI and ML systems that perform on a par with their black-box alternatives~\cite{chen2018interpretable}.%

This anecdotal trade-off and a tendency to prioritise predictive power mean that explainability is often only an afterthought. %
Such a mindset %
contributes to a landscape with an abundance of well-performing but inherently opaque algorithms that are in need of explainability, thus creating a demand for universal explainers that are post-hoc and model-agnostic, such as surrogates~\cite{craven1996extracting,ribeiro2016why,sokol2019blimey}. %
This seemingly uncompromising development approach -- where state-of-the-art performance remains the main objective, later complemented with a post-hoc explainer -- offers an attractive alternative (and rebuttal) to designing inherently explainable AI and ML systems, whose creation arguably requires more effort. %
While such explainers are compatible with any black-box model, they are not necessarily equally well suited for every one of them~\cite{sokol2020tut} -- after all the computer science folklore of ``no free lunch'' (a single, universal algorithm cannot outperform all the others across the board) applies here as well -- which is reflected in the continuous stream of novel XAI and IML techniques being proposed in the literature, many of whom report competing or contradictory findings (likely because of diverging operational contexts). %
Some post-hoc and model-agnostic explainers boast appealing properties and guarantees, however upon closer inspection one often encounters caveats and assumptions required for these to hold, such as the underlying ``black box'' being a linear model~\cite{lundberg2017unified}. %
Additionally, making an explainer model-agnostic introduces an extra layer of complexity that usually entails a degree of randomness and decreased fidelity~\cite{zhang2019should,sokol2019blimey}, so that using them may become a stopgap to claim explainability of an inherently opaque predictive system instead of addressing genuine explainability needs. %
Correctly interpreting the insights produced by XAI and IML methods may therefore be challenging as it requires a sufficient level of (technical) expertise and alignment with the explainees' background knowledge for the recipients to understand the capabilities and limitations of the explainer, thus avoid drawing incorrect conclusions -- %
this is especially relevant to post-hoc and model-agnostic approaches given their higher complexity~\cite{mittelstadt2019explaining,sokol2020limetree,sokol2020towards,sokol2020tut}.%

In \citeauthor{rudin2019stop}'s view~\cite{rudin2019stop}, many high-stakes AI and ML systems can be made explainable by design with enough effort put towards data pre-processing, feature engineering and modelling (which otherwise, e.g., for neural networks, may go into architecture search and parameter tuning). %
Such ante-hoc explainers are usually domain-specific and after the initial engineering endeavour they are easy to manage and maintain. %
While this approach should be championed for structured (tabular) data where it has been shown to perform on a par with state-of-the-art black boxes~\cite{chen2018interpretable}, the same may be unachievable for sensory data such as images and sounds, for which opaque models, e.g., deep neural networks, have the upper hand. %
In addition to black boxes modelling sensory data, pre-existing, inaccessible or legacy predictive systems may require interpretability, in which case they can only be retrofitted with post-hoc explainers. %
Such techniques are also helpful to engineers and developers working with predictive models since they enable inspection and debugging of data-driven systems. %
However, falling back on off-the-shelf solutions may not guarantee acceptable fidelity~\cite{sokol2019blimey,sokol2020limetree,sokol2020tut} (specifically, soundness and completeness), which is of particular importance and may require tailor-made explainers and transparent communication of their limitations.%

While composing a predictive pipeline, we have an abundance of pre-processing as well as modelling tools and techniques at our disposal, a selection of which will end up in the final system. %
The XAI and IML landscape, on the other hand, is quite different, especially for post-hoc and model-agnostic approaches: explainers tend to be end-to-end tools with only a handful of parameters exposed to the user. %
In view of ``no free lunch'', this is undesirable as despite being model-agnostic, i.e., compatible with any model type, these monolithic algorithms cannot perform equally well for every one of them~\cite{sokol2019blimey,sokol2020tut}. %
This variability in their behaviour can often be attributed to a misalignment between the assumptions baked into an explainer and the properties of the explained system, which manifests itself in low fidelity.%

Model-specific or ante-hoc explainers as advocated by \citet{rudin2019stop} can be used to address this issue; %
however, as discussed earlier, such a solution may have limited applicability and cannot be retrofitted to pre-existing predictive systems. %
Resolving a similar challenge in machine learning and data mining often comes down to a series of investigative steps to guide algorithmic choices down the line, which can be operationalised within a standardised process for knowledge discovery such as KDD~\cite{fayyad1996data}, CRISP-DM~\cite{chapman2000crisp,martinez2019crisp} or BigData~\cite{agrawal2012challenges}. %
For example, by analysing feature correlation, data uniformity and class imbalance, we can account for these phenomena when engineering features and training models, thereby making the resulting systems more accountable and robust. %
Nonetheless, while we may have a set of universal properties expected of XAI and IML systems~\cite{sokol2020explainability}, we lack a dedicated process that could guide the development and assessment of explainers -- their practical requirements and needs -- which likely hinders adherence to best practice. %
Although one can imagine a generic workflow for designing inherently interpretable (ante-hoc) systems~\cite{rudin2019stop}, a similar endeavour should not be neglected for model-agnostic and post-hoc explainers that could be adapted to individual predictive black boxes by capitalising on their flexibility and modularity~\cite{sokol2019blimey}, possibly overcoming low fidelity~\cite{sokol2020limetree}.%

More recently, in response to \citeauthor{rudin2019stop}'s call to action~\cite{rudin2019stop}, the disputed trade-off between \emph{transparency} and \emph{predictive power} has been revisited with a greater scientific rigour~\cite{herm2021don,bell2022its}. %
\citet{herm2021don} used explainability as a tool to aid people in problem-solving, investigating it under the aforementioned two dimensions complemented with \emph{comprehensibility}, which should be a direct consequence of a model being recognised as explainable. %
Their preliminary findings are somewhat at odds with \citeauthor{rudin2019stop}'s postulate, especially so for high-stakes scenarios for which the user studies suggest that an artificial neural network enhanced with SHAP explanations boasts high predictive performance and is also perceived as explainable. %
Similarly, \citet{bell2022its} performed empirical quantification of this trade-off in two public policy domains and found that (inherently) interpretable models may not necessarily be more explainable than black boxes. %
The results of their experiments show that even opaque models may be recognised as explainable -- nonetheless, the authors emphasise the importance of ante-hoc explainability in mission-critical domains -- hinting that the trade-off may be more nuanced than acknowledged in the literature. %
Notably, while the explainee's perception may suggest that black boxes accompanied by post-hoc explainers are up to the task, the fidelity, correctness and veracity of such insights remain contestable, especially in view of the recipient's susceptibility to be convinced by sheer presence of ``explanations'' that themselves may not necessarily be truthful~\cite{herman2017promise} or meaningful (as famously shown by The Copy Machine study~\cite{langer1978mindlessness}). %
With just a few such inquiries available, the topic requires further investigation to offer a clear view on the possible trade-off between transparency and predictive power in XAI and IML. %
We postulate that -- in accordance with our definition -- particular focus should be put on the \emph{reasoning} employed by explainees to extract meaning and create understanding based on transparent insights into black boxes given that a misconception of how to interpret them may result in a false sense of explainability~\cite{mittelstadt2019explaining}.%

A distinct viewpoint on this matter manifests itself in %
claims that we should not expect machine learning algorithms, such as deep neural networks, to be explainable and instead regulate them purely based on their real-life performance~\cite{simonite2019google} and behaviour~\cite[minute 29]{norvig2017google}, however it is not a widely shared belief~\cite{jones2018geoff}. %
This insight comes from the alleged inability of humans to explain their actions since such justifications are post-factum stories that are concocted and retrofitted for the benefit of the audience. %
Certifying autonomous agents based on their output, on the other hand, is consistent with human values as one can hypothesise about committing a crime, but one cannot be punished unless such a thought is acted upon. %
While the origin and nature of human thought processes may be shrouded in mystery, its formulation is expected to follow the reason of logic to be (socially) acceptable. %
In particular, \citet{miller2019but} refutes performance-based validation by arguing that explainability stemming from regulatory requirements is secondary to concerns arising from societal values such as ethics and trust. %
Importantly, making data-driven systems understandable can instil confidence into the public as it allows the creators of such technologies to justify, control and improve them as well as lead to new discoveries~\cite{adadi2018peeking}. %
An appropriate and comprehensive explainability solution can also become a technological springboard to reducing or eliminating bias~\cite{saxena2019perceptions,saxena2019fairness}, unfairness~\cite{buolamwini2018gender,olteanu2019social,kusner2017counterfactual} and lack of accountability (to the benefit of robustness~\cite{ackerman2019three,goodfellow2015explaining}, safety~\cite{angwin2016machine,grzywaczewski2017training} and security) from data-driven predictive models, thus improving their trustworthiness~\cite{sokol2019fairness}.%

\begin{figure}[!t]
    \centering
    \includegraphics[trim={6pt 87pt 6pt 87pt},clip,width=.65\textwidth]{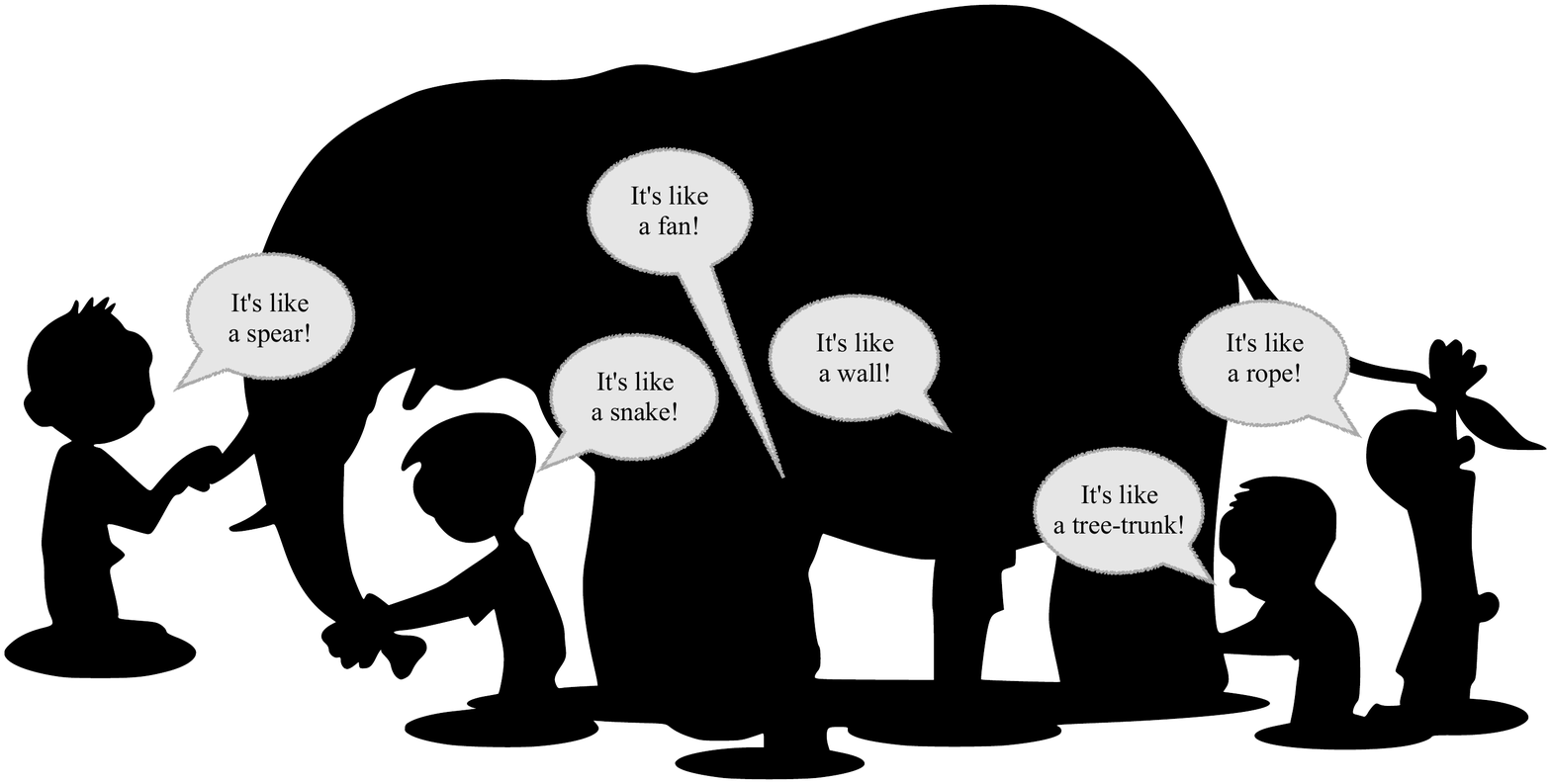}%
    \caption{%
Depiction of The Blind Men and the Elephant parable~\cite{saxe2016blind} illustrating that any complex subject can be studied in many ways. %
It also symbolises that individual pieces of evidence may often be contradictory and insufficient to understand the bigger picture without first being aggregated and grounded within a shared context.\label{fig:ch1:elephant}}%
\end{figure}

\section{Explanation Diversity and Multiplicity -- What to Explain and How to Explain It\label{sec:elephant}}

So far we have primarily focused on explaining predictions and actions of intelligent systems since they are observable and can be related to by a wide range of explainees regardless of their background. %
However, automated \textbf{predictions} are just artefacts of a more elaborate %
artificial intelligence or machine learning predictive process, which manipulates \textbf{data} to infer \textbf{models} that generalise well, thus are capable of predicting (previously unseen) instances~\cite{flach2012machine}. %
Since any element of this workflow can be opaque~\cite{sokol2017role,sokol2017roleARW}, %
comprehensive explanations may need to consist of insights pertaining to the entire predictive pipeline, discussing diverse topics such as data collection and (pre-)processing, modelling caveats and assumptions, and the meaning and interpretation of predictions, %
all of which can be %
bundled together in a shared user interface to provide a multi-faceted view of the investigated system~\cite{krause2016interacting,krause2016using,weld2019challenge}.%

Additionally, as each explanation may provide just a small, and quite possibly distorted, reflection of the true behaviour of a data-driven model, achieving the desired level of transparency (and understanding) might require communicating multiple, complementary insights for each unintelligible step or observation, which in turn bears the danger of overwhelming and confusing the explainee. %
This multitude of explanatory information has to be navigated carefully and can be understood as unique probing and inspection techniques that without a shared context may yield competing or even contradictory evidence akin to the parable of The Blind Men and the Elephant~\cite{saxe2016blind} %
illustrated in Figure~\ref{fig:ch1:elephant}. %
Note that this phenomenon is not unique to explainability; %
multiplicity of data-driven models all of whom exhibit comparable predictive performance despite intrinsic differences, sometimes called the Rashomon effect of statistics, is well documented~\cite{breiman2001statistical,marx2020predictive,fisher2019all,sokol2022ethical}. %
Furthermore, as AI and ML processes are directional -- from data, through models, to predictions -- the latter components depend on the former, which also applies to their respective explanations. %
For example, if data attributes are incomprehensible, explanations of models and predictions expressed in terms of these features will also be opaque.%

\textbf{Explaining data} may be challenging without any modelling assumptions, hence %
there may not necessarily exist a pure data explanation method beyond simple \emph{summary statistics} (e.g., class ratio or per-class feature distribution) and \emph{descriptors} (e.g., ``the classes are balanced'', ``the data are bimodal'' or ``these features are highly correlated''). %
Note that the former simply state well-defined properties and may not be considered explanations, whereas the latter can be contrastive and lead to understanding. %
Importantly, data are already a model -- they express a (subjective and partial) view of a phenomenon and come with certain assumptions, measurement errors or even embedded cultural biases (e.g., ``How much is a lot?''). %
Data statements~\cite{bender2018data}, data sheets~\cite{gebru2018datasheets} and nutrition labels~\cite{holland2018dataset} attempt to address such concerns by capturing these (often implicit) assumptions. %
As a form of data explanations, they characterise important aspects of data and their collection process in a coherent format, e.g., experimental setup, collection methodology (by whom and for what purpose), pre-processing (cleaning and aggregation), privacy aspects, data ownership, and so on.%

\textbf{Explaining models} %
in whole or in parts (e.g., specific sub-spaces or cohorts) %
should engender a general, truthful and accurate understanding of their functioning. %
While some predictive systems may be inherently transparent, e.g., shallow decision trees, their simulatability~\cite{lipton2016mythos} -- the explainee's ability to simulate their decision process mentally \emph{in vivo} -- may not produce understanding (see Section~\ref{sec:bbnes}). %
Popular model explanations include feature importance~\cite{breiman2001random,fisher2019all}, feature influence on predictions~\cite{friedman2001greedy}, presenting the model in cognitively-digestible portions~\cite{krause2016using,smilkov2016embedding} and model simplification~\cite{craven1996extracting} (e.g., mimicking its behaviour or a global surrogate). %
Since not all models operate directly on the input features, an \emph{interpretable representation} may be necessary to convey an explanation, e.g., a super-pixel segmentation of an image~\cite{ribeiro2016why}; alternatively, if the data are comprehensible, landmark exemplars can be used to explain the behaviour of a model or its parts~\cite{kim2014bayesian,kim2016examples}.%

\begin{wrapfigure}[15]{i}{.43\textwidth}%
    \centering
    \vspace{-.5\baselineskip}
    \includegraphics[width=.4\textwidth]{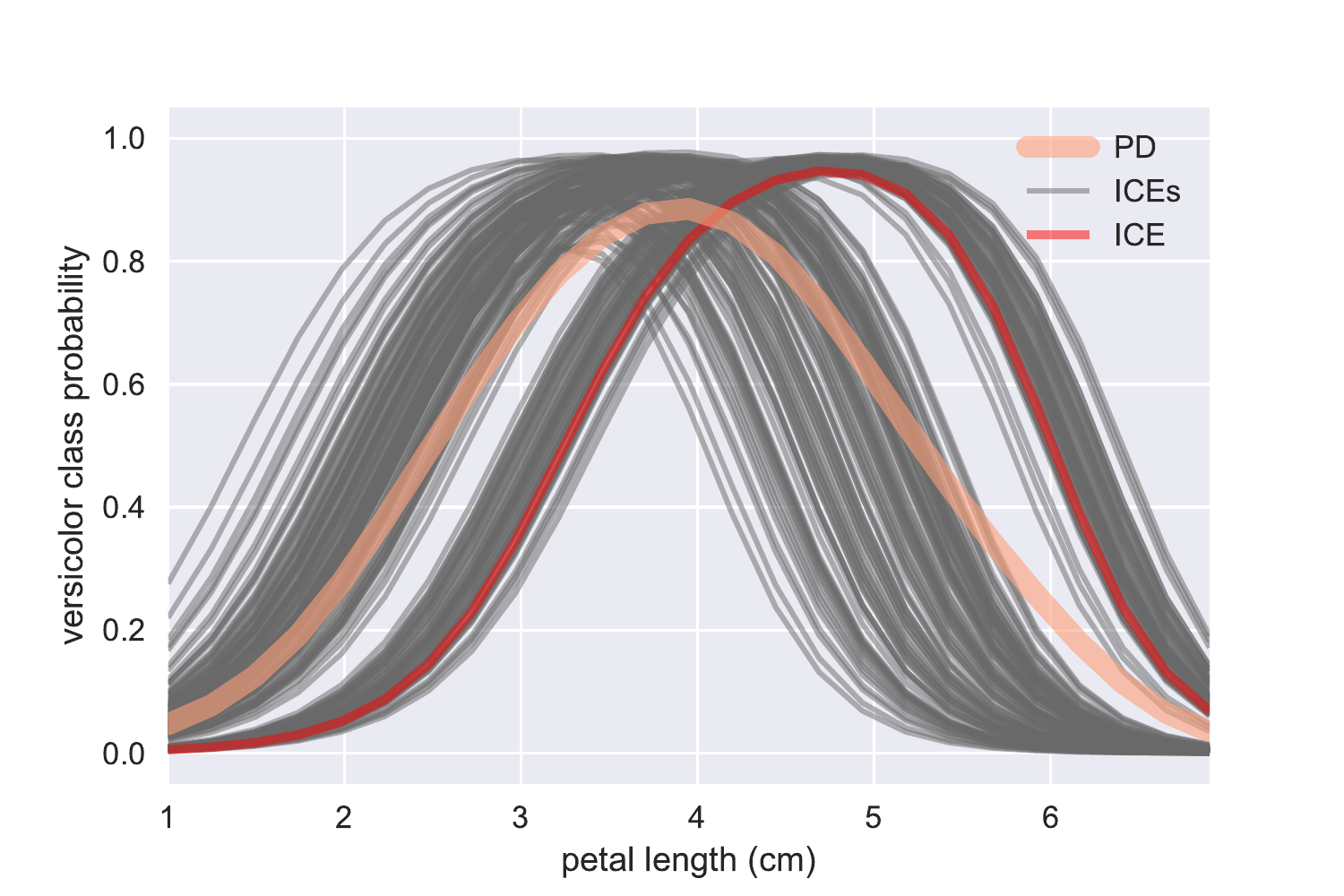}%
    \caption{%
Explanation of a model predicting the probability of the \emph{versicolor} class when varying the \emph{petal length} attribute for the Iris data set~\cite{fisher1936use}. %
Individual Conditional Expectation of a selected instance is plotted in red; %
the orange curve is the Partial Dependence of the model computed by averaging all individual ICEs (displayed in grey).%
\label{fig:ch1:ice-pd}}%
\end{wrapfigure}
\textbf{Predictions} are explained to communicate a rationale behind a particular decision of a model. %
Depending on the explanation type, a range of diverse aspects concerning the model's decisive process can be provided to the explainee. %
For example, the user may be interested in feature importance~\cite{ribeiro2016why}, feature influence~\cite{lundberg2017unified}, relevant data examples~\cite{kim2015ibcm} and training instances~\cite{koh2017understanding}, or contrastive statements~\cite{wachter2017counterfactual,poyiadzi2020face}, to name a few. %
Note that while some of these explanation types are similar to model explanations, here they are explicitly generated with respect to a single data point and may not necessarily generalise beyond this particular case, whereas for model explanations they convey similar information for all data (i.e., the entire modelled space). %
A good example of this duality is information communicated by Individual Conditional Expectation (ICE~\cite{goldstein2015peeking}) and Partial Dependence (PD~\cite{friedman2001greedy}), both of which are feature influence explanations -- the first with respect to a single data point and the latter concerning a model -- as shown in Figure~\ref{fig:ch1:ice-pd}. %
Akin to model explanations, the information can be conveyed in the raw feature space or using an interpretable representation.%

With such a diverse range (and possibly large quantity) of explanations, their presentation requirements -- \textbf{content}, \textbf{delivery format}, \textbf{communication medium} and \textbf{provision protocol} or \textbf{mechanism}~\cite{sokol2020explainability,gregor1999explanations} -- will naturally vary~\cite{sokol2017role,sokol2017roleARW}. %
A simple approach to characterise an AI component is (statistical) \emph{summarisation} -- it is commonly used for describing properties of data with numerical tables and vectors, which can be difficult to digest for non-experts. %
\emph{Visualisation} -- a graphical representation of a phenomenon -- is a more advanced, insightful and flexible analytical tool. %
Static figures communicate information in one direction, akin to summarisation; however, creating interactive plots can facilitate a ``dialogue'' with an explainee, thereby catering to a more diverse audience. %
Visualisations are often supported by short narratives in the form of captions, which increase their informativeness. %
\emph{Textualisation} -- a natural language description of a phenomenon -- can express concepts of higher complexity and dimensionality than plots, which can help to overcome the curse of dimensionality and the inherent limitations of the human visual system. %
Communicating with text enables a true dialogue and has been shown to be more insightful and effective than presenting raw, numerical and visual data~\cite{portet2009automatic}, which can accompany the narrative to improve its expressiveness. %
A further refinement of textualisation is formal \emph{argumentation}~\cite{dung2009assumption} -- a structured and logically-coherent dialogue accounting for every disputable statement and giving the explainee an opportunity to contest the narrative, thus providing explanations leading to understanding rather than informative descriptions. %
Finally, such explanatory processes can either be triggered automatically, invoked (and driven) by the users or offered contextually whenever a need for a clarification is detected by the explainer~\cite{gregor1999explanations}.%

Thus far we have been mainly concerned with AI and ML explainability on a relatively abstract level, all of which constitute just a small portion of XAI and IML research. %
In an ideal world, relevant publications would consider many of the aforementioned factors and build their mechanisms around them, however it has only recently become a trend and numerous early pieces of work lack such a reflection. %
To complement the viewpoint presented in the preceding sections and bridge the \emph{theoretical} (foundational and social) and \emph{technical} (algorithmic and engineering) aspects of explainers we briefly traverse through \textbf{practical explainability research}. %
Without aiming to be exhaustive -- given the availability of several comprehensive surveys~\cite{guidotti2018survey,linardatos2021explainable} -- we finish this section by identifying a number of landmark contributions that have influenced the entire research field. %
We also omit topics adjacent to explainability, such as interactive exploratory user interfaces~\cite{wexler2017facets,tensorboard}, creative visualisations of explainability approaches~\cite{krause2016interacting} and systems combining multiple explainability techniques within a single tool~\cite{weld2019challenge}.%

The most popular explainers are \emph{model-agnostic} and \emph{post-hoc} as they can be retrofitted into any predictive system (at the expense of adding a modelling layer that may negatively impact explanation fidelity). %
These include RuleFit~\cite{friedman2008predictive}, Local Interpretable Model-agnostic Explanations (LIME~\cite{ribeiro2016why}), anchors~\cite{ribeiro2018anchors}, SHapley Additive exPlanations (SHAP~\cite{lundberg2017unified}), Black-box Explanations through Transparent Approximations (BETA~\cite{lakkaraju2016interpretable,lakkaraju2017approximations}), PD~\cite{friedman2001greedy}, ICE~\cite{goldstein2015peeking} and Permutation Importance (PI~\cite{breiman2001random}), among many others. %
Most of these methods operate directly on raw data, with the exception of LIME and anchors, which use interpretable representations to improve intelligibility of explanations composed for complex data domains such as text and images. %
Another attractive avenue of explainability research, which partly overlaps with post-hoc methods, is opening up (deep) neural networks by designing tools and techniques \emph{specific to these approaches} or, more broadly, compatible with differentiable predictors. %
These models are notoriously opaque, however their superior predictive performance for a wide spectrum of applications increases their popularity and accelerates their proliferation~\cite{lecun2015deep}. %
Relevant explainability techniques include global surrogates~\cite{craven1996extracting}, saliency maps~\cite{zintgraf2017visualizing}, influential training instances~\cite{koh2017understanding}, counterfactuals~\cite{wachter2017counterfactual} (which are surprisingly similar to the problematic adversarial examples~\cite{goodfellow2015explaining}), and influential high-level, human-intelligible insights based on Testing with Concept Activation Vectors (TCAV~\cite{kim2018interpretability}). %
An alternative XAI and IML research agenda concentrates on \emph{inherently explainable} predictive models, and \emph{ante-hoc} explainers designed for popular data-driven systems. %
Examples of the former are generalised additive models~\cite{lou2013accurate} and falling rule list~\cite{wang2015falling}; whereas the latter include global and local explanations of na\"ive Bayes classifiers~\cite{kulesza2015principles}, and clustering insights based on prominent exemplars and dominating features~\cite{kim2014bayesian}.%

\section{Towards Understanding Facilitated Through Intelligible and Robust Explainers\label{sec:summary}}%

In this paper we explored the relatively recent and still evolving domains of artificial intelligence explainability and machine learning interpretability. %
We introduced the main topics and provided the philosophical, theoretical and technical background needed to appreciate the depth and complexity of this research. %
In particular, we highlighted two different mental models: \emph{functional} -- enough understanding to operationalise a concept; and \emph{structural} -- in-depth, theoretical appreciation of underlying processes. %
We further argued that the former -- a shallow form of understanding -- aligns with The Chinese Room Argument~\cite{searle1980minds,penrose1989emperor} and the notion of simulatability~\cite{lipton2016mythos}. %
We also reviewed diverse notions of explainability, interpretability, transparency, intelligibility and many other terms that are often used interchangeably in the literature, and argued in favour of \emph{explainability}. %
We defined this concept as (logical) \emph{reasoning} applied to transparent XAI and IML insights interpreted under specific \emph{background knowledge} within a given context -- a process that engenders \emph{understanding} in explainees. %
We used these observations to challenge the popular view that decision trees are explainable just because they are transparent. %
Deep or wide trees lack interpretability, which can be restored by applying a suitable form of logical reasoning -- a prerequisite of explainability -- undertaken by either an algorithm or a human investigator.%

While the most visible aspect of XAI and IML research is the technology that enables it, explainees -- the recipients of such explanations who tend to be humans -- are just as important (and ought to be treated as first-class citizens) since their \emph{understanding} of the underlying predictive system and its behaviour determines the ultimate success of an explainer. %
We explored this topic by looking at human-centred explainability and various desiderata that this concept entails, in particular focusing on explicitly acknowledging presence of humans and projecting the explanations directly at them. %
To this end, we pursued important insights from the social sciences that prescribe how to adapt machine explainability to fulfil expectations of the explainees, hence achieve seamless explanatory interaction. %
The two crucial observations in this space are: (i) a preference for (meaningful) \emph{contrastive} explanations, which form the cornerstone of human-centred explainability; and (ii) facilitating an interactive, dialogue-like, bi-directional explanatory \emph{process} -- akin to a conversation -- as opposed to delivering a one-off ``take it or leave it'' explanation %
to ensure coherence with people's expectations regardless of their background knowledge and prior experience with this sort of technology. %
Notably, the explanation type and delivery medium should also be adapted to the circumstances. %
This is particularly important when the audience is diverse as one predefined type of an explanation may be insufficient since it is unlikely to address all the possible concerns, questions and unique perspectives. %
An XAI or IML explainer that communicates through contrastive explanations and provides the explainees with an opportunity to interactively customise and personalise them~\cite{madumal2019grounded} -- offering a chance to contest and rebut them in case of a disagreement -- should therefore be considered the gold standard~\cite{wachter2017counterfactual,miller2019explanation}.%

In addition to enhancing explainee satisfaction, operating within this purview has other, far-reaching benefits such as enabling algorithmic fairness evaluation, accountability assessment and debugging of predictive models. %
It is also compatible with all the elements of the artificial intelligence or machine learning workflow -- which consists of data, models and predictions -- as each of these components may be in need of interpretability. %
In view of a variety of explainability approaches, each operating in a unique way, we also looked at the disputed trade-off between explainability and predictive power, the existence of which has mostly been supported by anecdotal evidence thus far, albeit recent studies show that this dependency may be more nuanced than previously expected. %
We then connected this debate to the distinction between inherent (ante-hoc) and retrofitted (post-hoc) explainability: the former provides explanations of superior quality but requires extensive engineering effort to be built, whereas the latter is flexible and universal at the expense of fidelity. %
While the former may be shunned due to the required work, we argued that building trustworthy post-hoc explainers may be just as complicated and demand just as much commitment since these seemingly easy to use tools %
conceal a complex process governing their composition and influencing their quality behind the facade of universality~\cite{sokol2019blimey,sokol2020limetree,sokol2020towards,sokol2020tut}. %
This considerable effort required to set them up, therefore, illuminates a crucial question: Is it better to spend time and effort on configuring post-hoc explainers or instead invest these resources into building inherently explainable predictive models? %
Unsurprisingly, there is no definitive answer given the uniqueness of each individual case, e.g., legacy systems and predictors built from scratch.%

Regardless of the particular implementation and operationalisation details, explainers of automated decision-making systems should adopt and embody as many of these findings as possible to engender trust in data-driven predictive models. %
Since each explanation reveals just a fragment of the modelling process and only the right mixture of evidence can paint the full picture, XAI and IML approaches need to be responsive and adapt seamlessly to the users' requests and expectations. %
Such an engaging algorithmic interlocutor should build logically consistent narratives and serve more as a guide and a teacher than a facts reporter. %
To this end, we need to develop an explanatory process built on top of a system that enables logical reasoning between intelligent agents: human--machine or machine--machine. %
An appropriate foundation -- managing the dialogue as well as tracking and storing the evolving knowledge base of the involved parties -- should benefit and encourage an interdisciplinary research agenda drawing from multiple areas of computer and social sciences. %
In the end, nonetheless, the explainee needs to be a savvy interrogator, asking the right questions and firmly navigating the entire process to understand the behaviour of such data-driven oracles. %
After all, in Arthur C.\ Clarke's words: %
``Any sufficiently advanced technology is indistinguishable from magic.'' %
While this view may partially reflect a broader perception of artificial intelligence and machine learning applications, the work presented here reconciles XAI and IML research published to date to establish a solid foundation for addressing open questions in an effort to demystify predictive algorithms and harness their full potential. %
The logical next step in this pursuit is development of a comprehensive framework, flexible protocol and suite of (quantitative \& qualitative) metrics to meaningfully evaluate the quality and effectiveness of explainable AI and interpretable ML techniques, allowing us to choose the best possible solution for each unique problem.%

\renewcommand{\acksname}{Acknowledgements}
\begin{acks}
This research was partially supported by %
the TAILOR project, funded by EU Horizon 2020 research and innovation programme under GA No 952215; and %
the ARC Centre of Excellence for Automated Decision-Making and Society, funded by the Australian Government through the Australian Research Council (project number CE200100005).%
\end{acks}

\section*{Author Contributions}
Conceptualisation, K.S.; Methodology, K.S.; Investigation, K.S.; Writing -- Original Draft, K.S.; Writing -- Review \& Editing, K.S.\ and P.F.; Supervision, P.F.; Funding Acquisition, P.F.%

\section*{Declaration of Interests}
The authors declare no competing interests.%

\bibliographystyle{ACM-Reference-Format}
\bibliography{manuscript}

\end{document}